\documentclass[11pt]{article}
\usepackage[hyperref]{ccl2021-en}
\usepackage{times}
\usepackage{url}
\usepackage{latexsym}
\usepackage{fancyhdr}

\usepackage{array}
\usepackage{booktabs} 
\usepackage{multirow}

\usepackage{multicol}
\usepackage{CJKutf8}
\usepackage{algorithm}
\usepackage{algpseudocode}
\usepackage{amsmath}
\usepackage{makecell}
\usepackage{graphics}
\usepackage{epsfig}
\usepackage{enumitem}

\usepackage{enumerate}
\usepackage[explicit]{titlesec}

\usepackage{microtype}

\newcommand{\tabincell}[2]{\begin{tabular}{@{}#1@{}}#2\end{tabular}}

\setitemize[1]{itemsep=1pt,partopsep=0pt,parsep=1.2pt,topsep=2pt}

\title{Unifying Discourse Resources with Dependency Framework}

\author{Yi Cheng$^{1}$ \quad Sujian Li$^{1}$ \quad Yueyuan Li$^{2}$\\
$^1$MOE Key Lab of Computational Linguistics, School of EECS, Peking University\\
$^2$School of Electronic Information and Electrical Engineering, Shanghai Jiao Tong University\\
 {\tt \{yicheng, lisujian\}@pku.edu.cn, rowena\_lee@sjtu.edu.cn} \\
}

\date{}

\begin{document}
\maketitle
\begin{abstract}
For text-level discourse analysis, there are various discourse schemes but relatively few labeled data, because discourse research is still immature and it is labor-intensive to annotate the inner logic of a text. In this paper, we attempt to unify multiple Chinese discourse corpora under different annotation schemes with discourse dependency framework by designing semi-automatic methods to convert them into dependency structures. We also implement several benchmark dependency parsers and research on how they can leverage the unified data to improve performance.\footnote{The data is available at \url{https://github.com/PKU-TANGENT/UnifiedDep}.}
\end{abstract}

\vspace{1mm}
\section{Introduction}
\vspace{1mm}

Discourse parsing aims to 
construct the logical structure of a text and label relations between discourse units, as shown in Fig. \ref{fig:HIT-CDTBdep} and Fig. \ref{fig:CDTBexp}.
Various discourse corpora have been developed to promote the discourse parsing technique.
There exist multiple discourse schemes such as Rhetorical Structure Theory (RST) \cite{Carlson2001} and PDTB \cite{PDTB2008}, which act as guidelines of various discourse corpora.
At the same time, text-level discourse annotation is complicated and laborious, so the scale of a single discourse corpora is often much smaller compared with other NLP tasks.

One way to conquer 
data sparsity is to use different discourse corpora through multi-task learning.
One way to conquer the data-sparsity problem is to multi-task learning multiple corpora simultaneously.
Prior efforts 
on that mainly focus on discourse relation classification between two text spans, without considering the whole discourse structure \cite{multitask2016Liu,multitask2017Li},
since one principle of multi-task learning is that the tasks should be closely related, 
but the discourse structures of different corpora vary a lot, e.g. shallow predicate-arguments structure 
(e.g. the relations in Fig. \ref{fig:HIT-CDTBdep}) and deep tree structure (e.g. the CDTB tree in Fig. \ref{fig:CDTBexp}).

In this work, we explore another possible way to simultaneously leverage different corpora: we unify the existing discourse corpora under one same framework to form a much larger dataset.
Our choice for this unified framework is dependency discourse structure (DDS), where elementary discourse units (EDUs) are directly related to each other without consideration of intermediate text spans. 
Because as \newcite{CLdep} have argued, DDS is a very general discourse representation framework, and they adopt a dependency perspective to evaluate the English discourse corpora RST-DT \cite{Carlson2001} and the related parsing techniques.

Our work explore the feasibility of unifying discourse resources under dependency framework with consideration of the following two questions:
(\romannumeral1) How to convert other discourse structures into DDS?
(\romannumeral2) How to make the best use of the unified data to improve discourse parsing techniques? 

Oriented by the questions above, we unify three Chinese discourse corpora under dependency framework: HIT-CDTB \cite{HITCDTB}, SU-CDTB \cite{Li2014} and Sci-CDTB \cite{CY2019}.
HIT-CDTB adopts the predicate-argument structure similar to PDTB, with a connective as predicate and two text spans as arguments.
Following rhetorical structure theory (RST), SU-CDTB uses a hierarchical tree to represent the inner structure of each text, with EDUs as its leaves and connectives as  intermediate nodes. 
Sci-CDTB is a small-scale DDS corpus composed of 108 scientific abstracts. 
It is the only Chinese DDS corpus as far as we know.

The primary obstacle of unifying these corpora is inconsistency of the representation schemes, such as granularity of EDU and definition of relation types. 
Besides, the predicate-argument structure of HIT-CDTB leads to the problem that some discourse relations between adjacent text spans are absent, so that the information provided by the original annotation might be insufficient to form a complete dependency structure.
To tackle these problems, we redefine granularity of EDU, 
conduct mapping among different relation sets, 
and design semi-automatic methods to convert other discourse structures into DDS.
As the automatic part, we design the dependency tree transformation method for each corpora. 
As the manual part, we proofread all the segmentation of EDUs to follow the same definition, complement necessary information, and correct the transformed dependency trees.

Different from \newcite{CLdep} who only consider conversion between RST and DDS, we attempt to unify more discourse schemes into dependency framework and explore whether discourse parsing techniques can be promoted by the unified dataset. 
Here we implement several discourse dependency parsers and research through experiment on how they can leverage the unified data to improve performance. 
Then we give out our findings about how to make better use of the unified discourse data.

Contributions of this paper are summarized  as follows:
\begin{itemize}
    \item we propose to integrate the existing discourse resources under a unified framework;
    \item we design a unified DDS framework and convert three Chinese discourse corpora into dependency structure in a semi-automatic way and get a unified large-scale dataset;
    \item we implement several discourse dependency parsers and explore how the unified data can be leveraged to improve parsing performance.
\end{itemize}

\vspace{2mm}
\section{Background}
In this section we mainly introduce the three Chinese discourse corpora: HIT-CDTB, SU-CDTB and Sci-CDTB, which we use in this work.

\paragraph{HIT-CDTB} Borrowing the discourse scheme of  PDTB, HIT-CDTB adopts one-predicate two-arguments structure, where a connective serves as the predicate and two text spans as two arguments, as shown in Fig. \ref{fig:HIT-CDTBdep}. 
The connective can be either explicitly identified if it already exists in the original text, or implicitly added by annotators,  
while the arguments can be phrases, clauses, sentences, or sentence groups.
Each connective corresponds to a relation type.
In total, there are 4 coarse-grained types (i.e., \emph{temporal}  \emph{causal}, \emph{comparative} and \emph{extension}) and 22 fine-grained types (e.g., \emph{temporal} is further divided as \emph{synchronous} and \emph{asynchronous}).
The documents cover multiple domains, such as news, editorials and popular science articles.
When labeling each document, the discourse relations are labeled locally by only considering adjacent text spans, so a complete logical structure of the text may not be obtained.

\paragraph{SU-CDTB} Similar to RST, SU-CDTB represents the inner structure of a text with a hierarchical tree with EDUs as leaves, and uses connectives as  intermediate nodes to indicate rhetorical relations.
The whole text is first divided into several text spans which are recursively divided until getting EDUs.
In SU-CDTB, EDUs are segmented according to the punctuation marks.
The nucleus-satellite relation structure in RST is retained through arrows in the trees.  
Connectives in SU-CDTB are also given some relation attributes.
500 news documents from the Chinese Treebank \cite{DBLP:journals/nle/XueXCP05} are annotated. A discourse tree is constructed for each paragraph rather than the entire document, with an average of 4.5 EDUs per tree, which are relatively shallow.
Besides, the top-down constructed tree cannot cover some particular discourse structures, such as non-adjacent relations.
So parsers trained only with this corpus are not very likely to analyze more complex discourse logic.





\paragraph{Sci-CDTB} Sci-CDTB is a small Chinese discourse dependency corpus, where EDUs are directly connected with discourse relations without intermediate levels. 
Its definition of EDU refers to RST-DT, with some modifications based on the linguistic characteristics of Chinese.
The head of each EDU and the relations between them are annotated. 
Sci-CDTB is a small-scale corpus, only composed of 108 annotated scientific abstracts, so it is hard to support the training of a competitive discourse parser. It is the only Chinese DDS corpus to the best of our knowledge.

\vspace{1mm}
\section{Unifying Discourse Corpora}
\vspace{1mm}

In this section, we introduce how to convert the three discourse corpora  into one unified framework,
which mainly involves two aspects, i.e., EDU segmentation and dependency tree construction. 
Here, we adopt the EDU segmentation guideline of Sci-CDTB, which is similar to RST-DT \cite{Carlson2001}. 
Basic discourse units of SU-CDTB and HIT-CDTB are divided mainly based on punctuation marks, which is inconsistent with our EDU definition, so we 
manually re-segment their EDUs to ensure the same segmentation rules. 

As for dependency tree construction,
we should ensure correctness of both the tree structure and the relations between them.
In a dependency tree, 
each 
relation connects a \emph{head} EDU to a  \emph{dependent} EDU. Each EDU should have one and only one \emph{head}. 
As the three corpora adopt different relation sets (22 relation types in HIT-CDTB$_{\text{dep}}$, 18  in SU-CDTB$_{\text{dep}}$
\footnote{HIT-CDTB$_{\text{dep}}$ and SU-CDTB$_{\text{dep}}$ refer to the converted HIT-CDTB and SU-CDTB before relation mapping.}
and 26 in Sci-CDTB), 
during conversion we keep the relations unchanged for each corpus, and then map them into 17 predefined relation types, which are basically the same as the ones of SU-CDTB, except that relation \emph{example illustration} is merged into \emph{explanation}. 
It requires to be further investigated whether there is a more appropriate mapping, and there exists the possibility that these relation sets have inherent incompatibility. 

Experiments are conducted both before and after relation mapping.

\vspace{2mm}
\subsection{Conversion of HIT-CDTB}

\begin{figure}
    \centering
    \includegraphics[width=\textwidth]{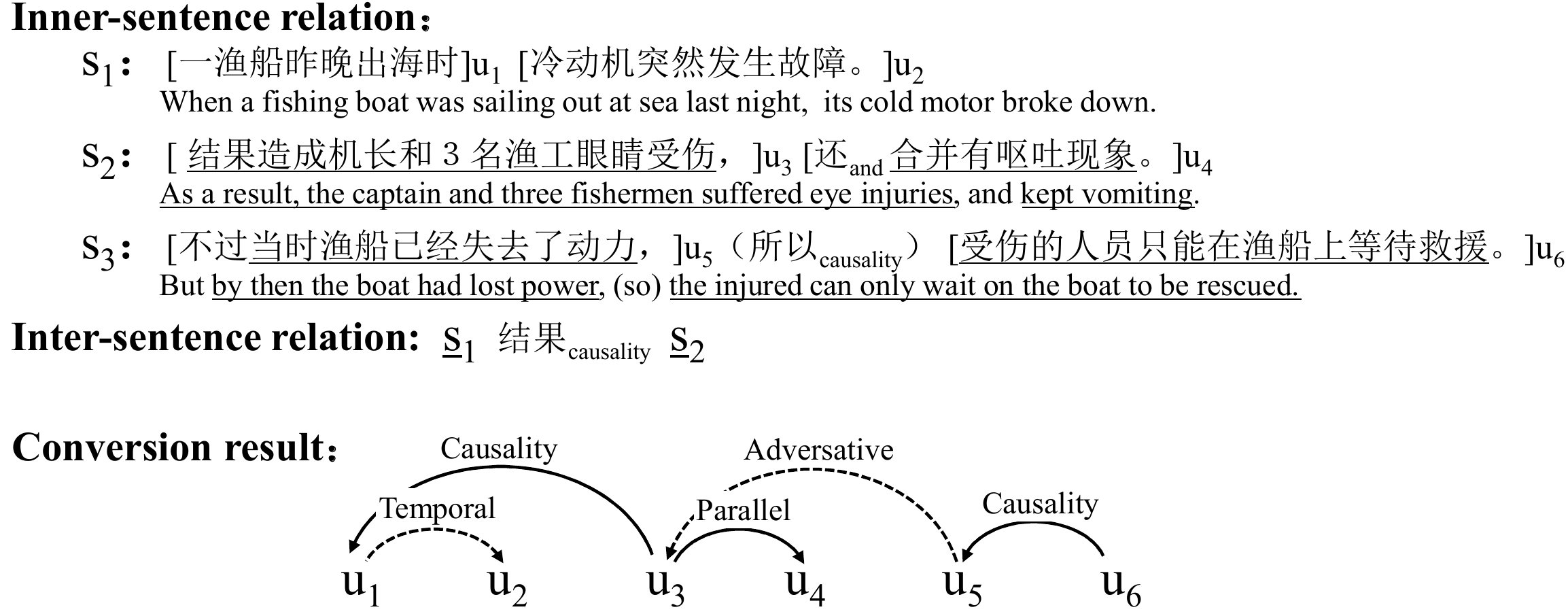}
  \caption{An example text in HIT-CDTB and its conversion result of DDS. Underlined texts are arguments of labeled HIT-CDTB discourse relations. The dependency relations derived from the original corpus are represented with solid lines in the conversion result, while the ones complemented during conversion are with dotted lines.
 }
 \vspace{1mm}
  \label{fig:HIT-CDTBdep}
\end{figure}
\begin{figure}
    \centering
    \includegraphics[width=\textwidth]{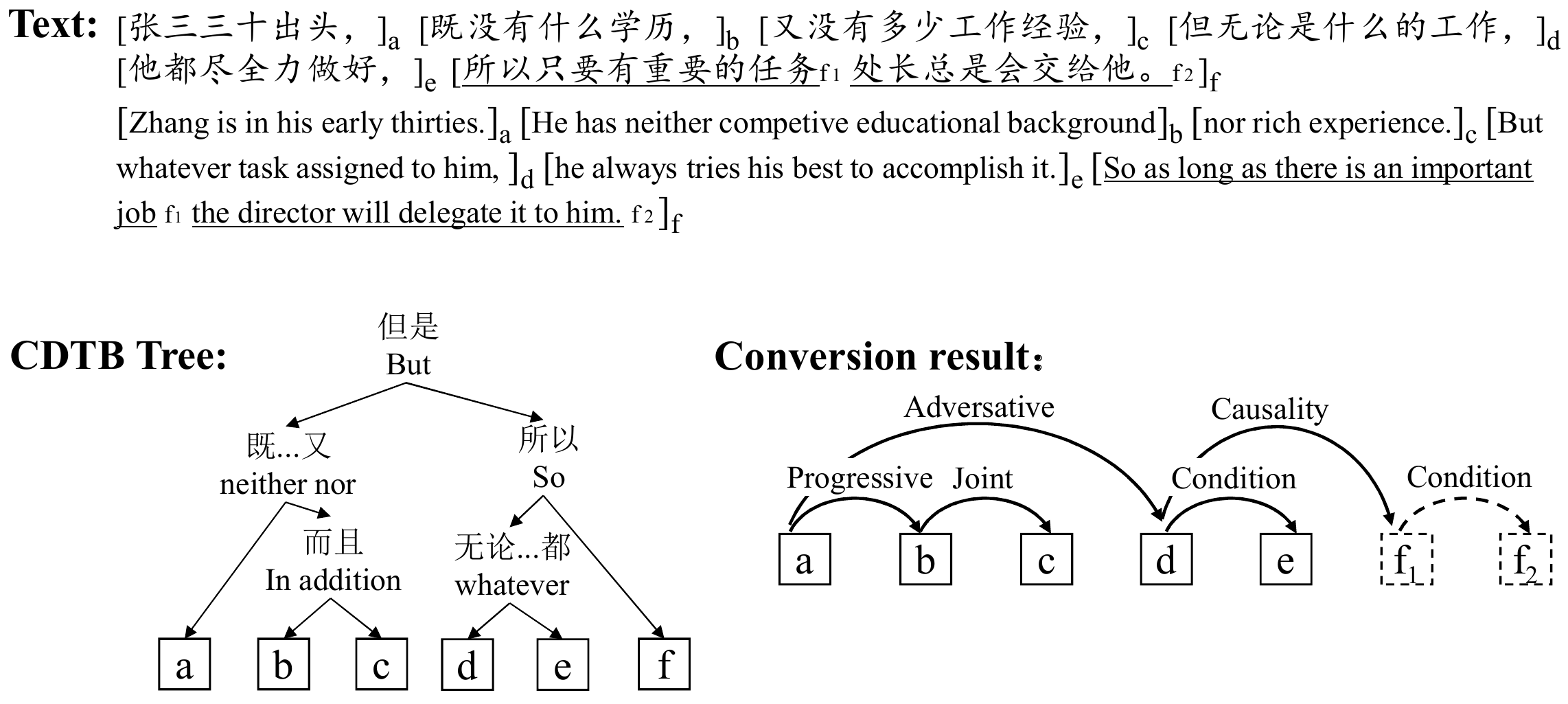}
  \caption{An example SU-CDTB tree and its conversion result of DDS }
  \label{fig:CDTBexp}
\end{figure}

Formally, a discourse relation in HIT-CDTB can be denoted as $R = <\overline{rel}_{HIT}, con,ARG1,ARG2>$, where $\overline{rel}_{HIT}$ is a relation type, $con$  a connective, and $ARG1$, $ARG2$ are two arguments. The arguments can be phrases, clauses, 
sentences or sentence groups. 
We will use the example in Fig. \ref{fig:HIT-CDTBdep} to elaborate the conversion process of HIT-CDTB.

For the example, HIT-CDTB labels two inner-sentence relations and one inter-sentence relation, which can not form a complete tree for the text.
As preparation for DDS construction, EDU segmentation is conducted with an pre-trained segmenter \cite{YangJf2018}, which are then manually checked to ensure quality.
With EDUs as basic units, we utilize the original relations labeled in HIT-CDTB and complement some relations to construct a complete dependency tree.
For instance, the relations derived from the original corpus and the ones complemented during conversion are respectively represented with solid lines and  dotted lines in the conversion result of Fig. \ref{fig:HIT-CDTBdep}.

Specifically, assume that $U_i$ is the EDU set covered by $Arg_i$ ($i=1,2$), the absent relations are \textbf{first} complemented to construct a complete dependency subtree. We generate these complemented relations in a rule-based way, by
summarizing common discourse markers for each relation type.  
For example, discourse markers  for \emph{Temporal} relation include \begin{CJK*}{UTF8}{gbsn} ``时''\end{CJK*}(when), \begin{CJK*}{UTF8}{gbsn} ``前''\end{CJK*}(before), \begin{CJK*}{UTF8}{gbsn} ``后''\end{CJK*}(after). 
\textbf{Then}, with two subtrees for $U_1$ and $U_2$, 
we replace $U_1$ and $U_2$ with their root EDUs as the arguments of $R'$ (e.g., \emph{Causality} relation between $u_1$ and $u_3$ in Fig. \ref{fig:HIT-CDTBdep}).
\textbf{Finally}, all the automatically-added relations are manually checked by two annotators. 

\vspace{1mm}
\subsection{Conversion of SU-CDTB}
\vspace{1mm}

Like RST, SU-CDTB represents the inner structure of a text with a hierarchical tree, with EDUs as its leaves and connectives as its intermediate nodes. The “nucleus-satellite” structure in RST is retained through the arrows on the tree. 
In SU-CDTB, EDU segmentation is conducted according to punctuation marks. 
Fig. \ref{fig:CDTBexp} shows an example discourse tree in SU-CDTB and its conversion result of DDS. 

The conversion process consists of three steps. 
\textbf{First}, as the annotation scheme of SU-CDTB is similar to that of RST-DT, we conduct conversion in a way similar to the algorithm of transformming RST-style discourse trees into DDS \cite{Li14,muller-etal-2012-constrained,hirao-etal-2013-single}. 
\textbf{Then}, we further subdivide some of the EDUs because EDU granularity in SU-CDTB is larger than our definition. \textbf{Finally}, we manually label the relations between newly-segmented EDUs, such as $f_1$ and $f_2$ in Fig. \ref{fig:CDTBexp}. As SU-CDTB constructs a complete tree for each text, the conversion process is relatively labor-saving. 

\vspace{1mm}
\subsection{Corpus Statistics}
\vspace{1mm}

Size of HIT-CDTB$_{\text{dep}}$, SU-CDTB$_{\text{dep}}$ and Sci-CDTB are presented in Table \ref{table:corpusSize}.
We also list the statistics of RST-DT and SciDTB \cite{Yang2018}, an English dependency corpus, for comparison.
We can see that Sci-CDTB has a much smaller scale than HIT-CDTB$_{\text{dep}}$ and SU-CDTB$_{\text{dep}}$. 
HIT-CDTB$_{\text{dep}}$ and SU-CDTB$_{\text{dep}}$ have similar number of relations, but the average length of each document (531.7 characters, 28.6 EDUs) in HIT-CDTB$_{\text{dep}}$ is larger than that in SU-CDTB$_{\text{dep}}$ (94.8 characters, 4.8 EDUs).
Size of the unified DDS corpus Unified$_{\text{dep}}$ is comparable to the two English discourse corpora.

Two annotators have participated in the manual labeling work. 
It takes them about 3 months and 3 weeks respectively to do the manual annotation and checking for HIT-CDTB and SU-CDTB. In comparison, it takes two annotators 3 months to build Sci-CDTB from scratch, which has a much smaller scale than the converted data, showing that converting existing resources is a relatively efficient way to generate dependency discourse dataset. 
Conversion of SU-CDTB is less time-consuming because its discourse structure is more complete than HIT-CDTB.
For HIT-CDTB, annotators take much more time on EDU segmentation and tree completion, since this corpus is constructed bottom up and its scheme is more different from our discourse dependency framework.
Table \ref{table:relation} shows distribution of the five most frequent relations in each corpus. We can see that the relation distribution in Sci-CDTB is quite different from the other corpora.

\begin{table}[t]
\centering
\begin{tabular}{c|cccc|cc}
\specialrule{1.2pt}{1pt}{1pt}
\textbf{{Corpus}}   & \textbf{HIT-CDTB}$_{\text{dep}}$ & \textbf{SU-CDTB}$_{\text{dep}}$ & \textbf{Sci-CDTB} & \textbf{Unified}$_{\text{dep}}$ & \textbf{SciDTB} & \textbf{RST-DT}\\ \specialrule{1.2pt}{0pt}{0.5pt}
\textbf{\#Doc} &  353 & 2332 & 108 & \textbf{2793} & 798 & 385 \\
\textbf{\#Rel} &  9796 & 8181 & 1392 & \textbf{19369} & 18978 & 21787 \\ 
\specialrule{1.2pt}{0.5pt}{0pt}
\end{tabular}
\label{table:corpusSize}
\caption{Corpus Size Comparison}
\vspace{2mm}
\end{table}

\begin{table}[t]
\centering
\begin{tabular}{c|cc|cc|cc|cc}
\specialrule{1.2pt}{1pt}{1pt}
\multirow{2}{*}{\bf {Corpus}} & \multicolumn{2}{c|}{\textbf{HIT-CDTB}$_{\text{dep}}$} & \multicolumn{2}{c|}{\textbf{SU-CDTB}$_{\text{dep}}$}     & \multicolumn{2}{c|}{\textbf{Sci-CDTB}}  & \multicolumn{2}{c}{\textbf{Unified}$_{\text{dep}}$} \\ \cline{2-9}
& Rel. & \% & {Rel.} & {\%} & {Rel.} & {\%} & {Rel.} & {\%}  \\ \specialrule{1.2pt}{0pt}{0.5pt}
\multirow{5}{*}{\bf \tabincell{c}{Relation\\Distribution}}& {joint}   & {46.4}               & {joint}           & {53.2}               & {elaboration}     & {29.3}   & {join}t & {52.7}            \\
& {explanation}  & {8.7}                & {explanation}    & {10.9}               & {joint}             & {17.0}      & {explanation} & {16.7}         \\
& {progressive}     &{ 8.5}                & {causality}       & {8.3}                & {enablement}        & {9.9}      & {causality} & {6.9}          \\
& {expression}    & {8.1 }               & {continuation}    & {6.8}               & {bg-general}        & {9.7}    & {continuation} & {4.3}            \\
& {causality}       & {6.2 }               & {goal}            & {4.1}                & {evaluation}        & {6.1}      & {progressive} & {4.1}          \\ \specialrule{1.2pt}{0.5pt}{0pt}    
\end{tabular}
\caption{Distribution of the 5 most frequently-used relations in HIT-CDTB$_{\text{dep}}$, {SU-CDTB}$_{\text{dep}}$ and Sci-CDTB}
\label{table:relation}
\end{table}

\vspace{1mm}
\section{Dependency Discourse Parsing}
\vspace{1mm}

\paragraph{Baselines} Following the work of \newcite{Yang2018}, we implement four Chinese discourse dependency parsers: 

\begin{itemize}
\item \textbf{Graph-based Parser} adopts Eisner algorithm to predict the most possible dependency tree structure \cite{Li2014}. It refers to the graph-based method of syntax parsing, which adopts Eisner algorithm and MST algorithm to predict the most possible dependency tree structure. For simplicity, an averaged perceptron is implemented to train weights.
\item \textbf{Vanilla Transition-based Parser} adopts the transition method of dependency parsing proposed by \newcite{Nivre2003}, employing the action set of arc-standard system \cite{Nivre2004}.
An SVM classifier is trained to predict transition action for a given configuration.
\item \textbf{Two-stage Transition-based Parser} \cite{Wang2017} first adopts the transition-based method to construct a naked dependency tree, and then uses another SVM to predict relations, which can take  the tree-structure as  features.
\item \textbf{Bert-based Parser} also conducts parsing in a two-stage transition way, but in the second stage it uses a bert-based model, which incorporates BERT \cite{bert} with one additional output layer, to identify relation types. It keeps the pre-trained parameters\footnote{https://github.com/huggingface/transformers} and is fine-tuned on our task using Adam. 
\end{itemize}

\begin{table}[t]
  \centering
     \makeatletter\def\@captype{table}\makeatother
        \begin{tabular}{cccc}
        \specialrule{1.2pt}{1pt}{1pt}
                 & \textbf{Train} & \textbf{Dev} & \textbf{Test} \\ \specialrule{1.2pt}{0pt}{0.5pt}
        HIT-CDTB$_{\text{dep}}$ & 250   & 50  & 53   \\
        SU-CDTB$_{\text{dep}}$     & 1600  & 400 & 332  \\
        Sci-CDTB  & 68    & 20  & 20   \\ \hline
        Unified$_{\text{dep}}$  & 1918  & 470 & 405  \\ 
        \specialrule{1.2pt}{0.5pt}{0pt}
        \end{tabular}
  \caption{Division of training, validation and test set for the three corpora}
  \vspace{2mm}
  \label{table:division}
\end{table}

\begin{table}[t]
\centering
\begin{tabular}{c|ccc|ccc|ccc}
\specialrule{1.2pt}{1pt}{1pt}
                     & \multicolumn{3}{c|}{\textbf{HIT-CDTB}$_{\text{dep}}$} & \multicolumn{3}{c|}{\textbf{SU-CDTB}$_{\text{dep}}$}& \multicolumn{3}{c}{\textbf{Sci-CDTB}} \\ 
                     & \textbf{UAS}     & \textbf{LAS}$_{\text{O}}$  & \textbf{LAS}$_{\text{U}}$  & \textbf{UAS}       & \textbf{LAS}$_{\text{O}}$   & \textbf{LAS}$_{\text{U}}$  & \textbf{UAS}       & \textbf{LAS}$_{\text{O}}$ & \textbf{LAS}$_{\text{U}}$ \\ \specialrule{1.2pt}{0pt}{0.5pt}
Graph-based          & 0.353            & 0.237     &  0.255     & 0.585              & 0.415      & 0.428    & 0.338              & 0.175    &  0.199 \\ 
Vanilla Trans   & \textbf{0.835}   & 0.551    & 0.588      & \textbf{0.803}    & 0.580       &  0.588   & \textbf{0.525}     & 0.276   & 0.299   \\ 
Two-stage & \textbf{0.835}   & \textbf{0.565} & \textbf{0.600} & \textbf{0.803}    & 0.587 & 0.597  & \textbf{0.525}   & 0.276 & 0.299\\
Bert-based         & \textbf{0.835} & 0.564  & 0.576 & \textbf{0.803}  & \textbf{0.767} & \textbf{0.783}  & \textbf{0.525} & \textbf{0.358}   & \textbf{0.408}  \\ \hline
Two-stage$_{\text{Uni}}$ &  0.813  & -  & \textbf{0.614}  & 0.802 & - & 0.591  & \textbf{0.603}   &  -    &   0.393 \\
Bert-based$_{\text{Uni}}$  &   0.813  &  -  & 0.606  & 0.802 & - & 0.637  &  \textbf{0.603}  &  -    & 0.220    \\ \hline
Human                & 0.872            & 0.723     &  -  & 0.897              & 0.774        & -   & 0.806              & 0.627 & -\\ \specialrule{1.2pt}{0.5pt}{0pt}
\end{tabular}
\caption{Performance Comparison of Benchmark Parsers on different dependency discourse corpora}
\label{table:performanceCmp}  
\end{table}

\paragraph{Metrics} As for metrics, we use UAS and LAS to measure the dependency tree labeling accuracy without and with relation labels respectively.
LAS$_{\text{O}}$ and LAS$_{\text{U}}$ denote using the original relation set of each corpus or the predefined unified relation set. 

\paragraph{Results} Table \ref{table:performanceCmp} compares performance of the benchmark parsers.
To give a rough upper bound of the parsing performance, the last row lists the consistency of human annotation  by comparing two annotators' labelling results on 30 documents from each corpus respectively.
Division of training, validation and test set is shown in Table \ref{table:division}. 
The first four methods in the first block
show the results of the benchmark parsers which are trained and tested respectively on HIT-CDTB$_{\text{dep}}$, SU-CDTB$_{\text{dep}}$ and Sci-CDTB. 

\emph{Graph-based} method is less effective than the others,  but it could probably be improved by using other training methods like MIRA.
Among the transition-based methods, \emph{Bert-based} performs the best on SU-CDTB$_{\text{dep}}$ and Sci-CDTB$_{\text{dep}}$ with respect to LAS, but is 2.4\% lower than \emph{Two-stage} on HIT-CDTB$_{\text{dep}}$, probably because HIT-CDTB$_{\text{dep}}$ is a multi-domain corpus, and feature-based methods are more robust to changes in different domains.
Comparing the three corpora, parsing results on Sci-CDTB are the worst because Sci-CDTB is too small for supervised learning. As we expect, SU-CDTB performs the best since it has relatively shallow tree structure and its labeling is highly consistent.
With original relations mapped to the unified relation set, LAS$_{\text{U}}$ results of all the methods on the three corpora have been improved compared with LAS$_{\text{O}}$, proving that the unified relation set we use is acceptable.
Two-stage transition-based method performs better than the vanilla one with respect to LAS due to the addition of tree structural features in relation type prediction.



Two-stage$_{\text{Uni}}$ and Bert-based$_{\text{Uni}}$ use Unified$_{\text{dep}}$
as training data, and are tested on each corpus respectively.
From Table \ref{table:performanceCmp}, we can see that Sci-CDTB has a significant promotion of 7.8\% on UAS, while HIT-CDTB$_{\text{dep}}$ and SU-CDTB$_{\text{dep}}$ slightly decline. 
One possible explanation is that documents in HIT-CDTB$_{\text{dep}}$ are much longer than in SU-CDTB$_{\text{dep}}$ so their structures are not similar enough to improve each other's performance.
For Sci-CDTB, however, it also has a small numbers of EDUs per text so parsing performance can be improved by learning the augmented data of short trees from SU-CDTB.

For LAS$_{\text{U}}$, both methods increase on HIT-CDTB$_{\text{dep}}$, but drop on SU-CDTB$_{\text{dep}}$. As HIT-CDTB$_{\text{dep}}$ covers multiple domains and has an average of 28.6 EDUs per text, its annotation is more difficult and the DDS-conversion process is also more complicated. In comparison, with an average of 4.5 EDUs per tree, SU-CDTB only focuses on news domain and its DDS-conversion is relatively easy. As a result, SU-CDTB$_{\text{dep}}$ keeps a better data consistency than HIT-CDTB$_{\text{dep}}$, which may explain why the augmented data promotes the results on HIT-CDTB$_{\text{dep}}$, but introduces noise data for SU-CDTB$_{\text{dep}}$.
As Sci-CDTB is from scientific domain and its relations are much different from those of HIT-CDTB and SU-CDTB, the augmented labeled relations do not bring much improvement on relation labeling.

Overall, Unified$_{\text{dep}}$ can serve as a discourse dataset for researching cross-domain or long text discourse parsing.
However, our unification method may also introduce noise due to difference in text length and domain,
so it remains to be considered how to better leverage this unified large corpus to improve the Chinese discourse parsing techniques.

\vspace{1mm}
\section{Conclusions}
\vspace{1mm}

In our work, we design semi-automatic methods to unify three Chinese discourse corpora with dependency framework. Our methods of converting PDTB-style and RST-style discourse annotation into DDS can be used to convert more other corpora and further enlarge our unified dataset.
At the same time, by implementing several benchmark parsers on the converted data, we find that augmenting training set with the unified data can to some extent improve performance, 
but may also introduce noise and bring performance 
loss due to difference in text length and domain. 
This unified dataset is potentially helpful to research on cross-domain and long text discourse parsing, which will be our future work.

\section*{Acknowledgments}
We thank the NLP Lab at Soochow University and HIT-SCIR for offering the discourse resources (\emph{i.e.} SU-CDTB and HIT-CDTB) to conduct the research.

\bibliographystyle{ccl}
\bibliography{ccl2021-en}

\begin{thebibliography}{}

\bibitem[\protect\citename{Carlson \bgroup et al.\egroup }2001]{Carlson2001}
Lynn Carlson, Daniel Marcu, and Mary~Ellen Okurovsky.
\newblock 2001.
\newblock Building a discourse-tagged corpus in the framework of rhetorical
  structure theory.
\newblock In {\em Proceedings of the {SIGDIAL} 2001 Workshop, The 2nd Annual
  Meeting of the Special Interest Group on Discourse and Dialogue, Saturday,
  September 1, 2001 to Sunday, September 2, 2001, Aalborg, Denmark}.

\bibitem[\protect\citename{Cheng and Li}2019]{CY2019}
Yi~Cheng and Sujian Li.
\newblock 2019.
\newblock Zero-shot {C}hinese discourse dependency parsing via cross-lingual
  mapping.
\newblock In {\em Proceedings of the 1st Workshop on Discourse Structure in
  Neural NLG}, pages 24--29, Tokyo, Japan, November. Association for
  Computational Linguistics.

\bibitem[\protect\citename{Devlin \bgroup et al.\egroup }2018]{bert}
Jacob Devlin, Ming{-}Wei Chang, Kenton Lee, and Kristina Toutanova.
\newblock 2018.
\newblock {BERT:} pre-training of deep bidirectional transformers for language
  understanding.
\newblock {\em CoRR}, abs/1810.04805.

\bibitem[\protect\citename{Hirao \bgroup et al.\egroup
  }2013]{hirao-etal-2013-single}
Tsutomu Hirao, Yasuhisa Yoshida, Masaaki Nishino, Norihito Yasuda, and Masaaki
  Nagata.
\newblock 2013.
\newblock Single-document summarization as a tree knapsack problem.
\newblock In {\em Proceedings of the 2013 Conference on Empirical Methods in
  Natural Language Processing}, pages 1515--1520, Seattle, Washington, USA,
  October. Association for Computational Linguistics.

\bibitem[\protect\citename{Li \bgroup et al.\egroup }2014a]{Li14}
Sujian Li, Liang Wang, Ziqiang Cao, and Wenjie Li.
\newblock 2014a.
\newblock Text-level discourse dependency parsing.
\newblock In {\em Proceedings of the 52nd Annual Meeting of the Association for
  Computational Linguistics, {ACL} 2014, June 22-27, 2014, Baltimore, MD, USA,
  Volume 1: Long Papers}, pages 25--35.

\bibitem[\protect\citename{Li \bgroup et al.\egroup }2014b]{Li2014}
Yancui Li, Wenhe Feng, Jing Sun, Fang Kong, and Guodong Zhou.
\newblock 2014b.
\newblock Building chinese discourse corpus with connective-driven dependency
  tree structure.
\newblock In {\em Proceedings of the 2014 Conference on Empirical Methods in
  Natural Language Processing, {EMNLP} 2014, October 25-29, 2014, Doha, Qatar,
  {A} meeting of SIGDAT, a Special Interest Group of the {ACL}}, pages
  2105--2114.

\bibitem[\protect\citename{Li \bgroup et al.\egroup }2017]{multitask2017Li}
Haoran Li, Jiajun Zhang, and Chengqing Zong.
\newblock 2017.
\newblock Implicit discourse relation recognition for english and chinese with
  multiview modeling and effective representation learning.
\newblock {\em {ACM} Trans. Asian Low Resour. Lang. Inf. Process.},
  16(3):19:1--19:21.

\bibitem[\protect\citename{Liu \bgroup et al.\egroup }2016]{multitask2016Liu}
Yang Liu, Sujian Li, Xiaodong Zhang, and Zhifang Sui.
\newblock 2016.
\newblock Implicit discourse relation classification via multi-task neural
  networks.
\newblock In Dale Schuurmans and Michael~P. Wellman, editors, {\em Proceedings
  of the Thirtieth {AAAI} Conference on Artificial Intelligence, February
  12-17, 2016, Phoenix, Arizona, {USA}}, pages 2750--2756. {AAAI} Press.

\bibitem[\protect\citename{Morey \bgroup et al.\egroup }2018]{CLdep}
Mathieu Morey, Philippe Muller, and Nicholas Asher.
\newblock 2018.
\newblock A dependency perspective on rst discourse parsing and evaluation.
\newblock {\em Computational Linguistics}, 44(2):197--235.

\bibitem[\protect\citename{Muller \bgroup et al.\egroup
  }2012]{muller-etal-2012-constrained}
Philippe Muller, Stergos Afantenos, Pascal Denis, and Nicholas Asher.
\newblock 2012.
\newblock Constrained decoding for text-level discourse parsing.
\newblock In {\em Proceedings of {COLING} 2012}, pages 1883--1900, Mumbai,
  India, December. The COLING 2012 Organizing Committee.

\bibitem[\protect\citename{Nivre \bgroup et al.\egroup }2004]{Nivre2004}
Joakim Nivre, Johan Hall, and Jens Nilsson.
\newblock 2004.
\newblock Memory-based dependency parsing.
\newblock In {\em Proceedings of the Eighth Conference on Computational Natural
  anguage Learning (CoNLL-2004) at HLT-NAACL 2004}, pages 444--449. Association
  for Computational Linguistics.

\bibitem[\protect\citename{Nivre}2003]{Nivre2003}
Joakim Nivre.
\newblock 2003.
\newblock An efficient algorithm for projective dependency parsing.
\newblock {\em Proceedings of the 8th International Workshop on Parsing
  Technologies (IWPT)}, 07.

\bibitem[\protect\citename{Prasad \bgroup et al.\egroup }2008]{PDTB2008}
Rashmi Prasad, Nikhil Dinesh, Alan Lee, Eleni Miltsakaki, Livio Robaldo,
  Aravind~K. Joshi, and Bonnie~L. Webber.
\newblock 2008.
\newblock The penn discourse treebank 2.0.
\newblock In {\em Proceedings of the International Conference on Language
  Resources and Evaluation, {LREC} 2008, 26 May - 1 June 2008, Marrakech,
  Morocco}.

\bibitem[\protect\citename{Wang \bgroup et al.\egroup }2017]{Wang2017}
Yizhong Wang, Sujian Li, and Houfeng Wang.
\newblock 2017.
\newblock A two-stage parsing method for text-level discourse analysis.
\newblock In {\em Meeting of the Association for Computational Linguistics}.

\bibitem[\protect\citename{Xue \bgroup et al.\egroup
  }2005]{DBLP:journals/nle/XueXCP05}
Naiwen Xue, Fei Xia, Fu{-}Dong Chiou, and Martha Palmer.
\newblock 2005.
\newblock The penn chinese treebank: Phrase structure annotation of a large
  corpus.
\newblock {\em Nat. Lang. Eng.}, 11(2):207--238.

\bibitem[\protect\citename{Yang and Li}2018a]{Yang2018}
An~Yang and Sujian Li.
\newblock 2018a.
\newblock Scidtb: Discourse dependency treebank for scientific abstracts.
\newblock In {\em Proceedings of the 56th Annual Meeting of the Association for
  Computational Linguistics (Volume 2: Short Papers)}, pages 444--449.
  Association for Computational Linguistics.

\bibitem[\protect\citename{Yang and Li}2018b]{YangJf2018}
Jingfeng Yang and Sujian Li.
\newblock 2018b.
\newblock Chinese discourse segmentation using bilingual discourse commonality.
\newblock In {\em https://arxiv.org/abs/1809.01497}. arXiv.

\bibitem[\protect\citename{Zhang \bgroup et al.\egroup }2014]{HITCDTB}
Muyu Zhang, Bing Qin, and Ting Liu.
\newblock 2014.
\newblock Chinese discourse relation semantic taxonomy and annotation.
\newblock {\em Journal of Chinese Information Processing}, 28:26--28.

\end{thebibliography}

\end{document}